\documentclass[10pt]{JournalTitle}
\usepackage{graphicx}
\usepackage[singlelinecheck=off,font={footnotesize,bf},labelsep=space]{caption}
\usepackage{float}
\usepackage{amsmath}
\usepackage{amssymb}
\usepackage{microtype}
\title{ART: Action-based Reasoning Task Benchmarking for Medical AI Agents}
\titleshort{}

\author{
	Ananya Mantravadi$^{1,*}$,
    Shivali Dalmia$^{1,*}$,
	Abhishek Mukherji$^{1}$}
\footnotetext[1]{$^{1}$Centific Global Solutions, Inc. $^{*}$Equal contribution}
\doi{}

\abstract{Reliable clinical decision support requires medical AI agents capable of safe, multi-step reasoning over structured electronic health records (EHRs). While large language models (LLMs) show promise in healthcare, existing benchmarks inadequately assess performance on action-based tasks involving threshold evaluation, temporal aggregation, and conditional logic. We introduce \textbf{ART}, an Action-based Reasoning clinical Task benchmark for medical AI agents, which mines real-world EHR data to create challenging tasks targeting known reasoning weaknesses. Through analysis of existing benchmarks, we identify three dominant error categories: retrieval failures, aggregation errors, and conditional logic misjudgments. Our four-stage pipeline—scenario identification, task generation, quality audit, and evaluation—produces diverse, clinically validated tasks grounded in real patient data. Evaluating GPT-4o-mini and Claude 3.5 Sonnet on 600 tasks shows near-perfect retrieval after prompt refinement, but substantial gaps in aggregation (28-64\%) and threshold reasoning (32-38\%). By exposing failure modes in action-oriented EHR reasoning, ART advances toward more reliable clinical agents, an essential step for AI systems that reduce cognitive load and administrative burden, supporting workforce capacity in high-demand care settings}


\begin{document}

    \maketitle
    \begin{multicols}{2}

    \section{Introduction}

A global healthcare worker shortage of at least ten million is projected by 2030, threatening access to essential care and demanding interventions that can grow, retain, and empower the clinical workforce \cite{kumar2025heartbeat}. Closing this gap is not merely a staffing goal — eliminating the shortage could reduce the global disease burden by 7\%, avert 189 million years of life lost, and generate an estimated \$1.1 trillion in economic value.

One promising pathway toward scaling clinical capacity is the integration of artificial intelligence into clinical decision support systems, where automated reasoning over electronic health records (EHRs) can reduce cognitive load and accelerate high-frequency decision tasks \cite{Singhal25-expertmedicalQnA}. However, deploying such agents in medical settings requires reliability far beyond typical LLM benchmarks, particularly for tasks requiring multi-step reasoning (e.g., retrieving a lab value, checking whether it is out of range, then ordering a lab referral), threshold evaluation (e.g., assessing whether potassium replacement is indicated), and action-based decision-making (e.g., ordering a medication or triggering follow-up testing). 

Large language models (LLMs) have demonstrated strong performance in medical question answering and information retrieval~\cite{R}. Yet their reliability on these types of action-oriented clinical tasks remains largely understudied.
 Existing benchmarks, including Stanford's MedAgentBench~\cite{R}, primarily emphasize query-based reasoning (e.g., retrieving lab values or patient history) and leave critical gaps in systematic evaluation of procedural decision-making in dynamic clinical environments. Errors in clinical reasoning, such as misinterpreting laboratory values, incorrectly aggregating temporal data, or failing to apply treatment thresholds, are not benign: they directly translate into patient-level clinical risk. Our research is motivated by this increasing pressure on global healthcare systems, especially as populations age and care complexity rises. We aim to evaluate and strengthen the robustness of AI agents for healthcare and wellbeing services as a path to reducing caregiver workload, stress, and downstream economic burden.

\subsection{Limitations of Current Evaluation Approaches}

Despite recent progress, several fundamental challenges constrain the reliability of medical agent assessment:

\begin{itemize}
    \item \textbf{Limited coverage of action-based tasks}: Few datasets capture the complete reasoning chain from data retrieval through condition-based action execution.
    
    \item \textbf{Underexamined reasoning errors}: Threshold misinterpretations, temporal boundary errors, and conditional logic gaps remain underexamined in current benchmarks.
    
    \item \textbf{Lack of failure-targeted task generation}: Current benchmarks do not contain tasks that target known reasoning weaknesses.
\end{itemize}

\subsection{Our Contributions}

To address these limitations, we introduce \textbf{ART} (Action-based reasoning task benchmarking), a systematic framework for creating clinically grounded benchmark tasks that expose and evaluate reasoning weaknesses in medical AI agents. Our key contributions include:

\begin{itemize}
    \item \textbf{Systematic failure mode analysis}: We characterize three dominant failure types in medical agent tasks (retrieval, aggregation, multi step conditional logic) and use these to guide task construction.

    \item \textbf{Agentic task generation pipeline}: A four-stage workflow combining EHR failure pattern mining, LLM-based generation, and clinician review to produce clinically grounded tasks. Human-in-the-loop audit ensures task realism and clinical fidelity.

    \item \textbf{Expanded reasoning coverage at scale}: We generate 600{+} structured tasks across 11 lab codes \cite{loinc2024}, targeting temporal aggregation and threshold-based reasoning that current benchmarks lack.

    \item \textbf{Empirical evaluation of state-of-the-art models}: GPT-4o-mini\cite{gpt_4o} and Claude-3.5 Sonnet \cite{anthropic2024claude35sonnet} show near-perfect retrieval accuracy but performance degrades for aggregation (28–64\%) and conditional reasoning (32–38\%).
\end{itemize}

Our framework generates synthetic tasks grounded in real EHR-formatted patient data, ensuring clinical realism while enabling systematic evaluation at scale. By targeting known failure modes and incorporating clinical expert validation, ART provides a foundation for continuous improvement of medical AI agent reasoning capabilities. Our implementation interfaces with Fast Healthcare Interoperability Resources \cite{hl7fhir2024} (FHIR) APIs to align with the standardized access layer used in modern EHR systems.

\section{Failure Mode analysis}

To identify bottlenecks in clinical agent reasoning, we conducted a systematic failure analysis on the MedAgentBench dataset, focusing on tasks involving patient's information retrieval, temporal aggregation, and threshold based conditional decision-making. Each agent output was compared against gold standard tasks to categorize the underlying error patterns. Three major failure categories were identified: \textit{Data Retrieval Failure}, \textit{Aggregation Error}, and \textit{Threshold based conditional misjudgement} (see Table \ref{tab:failure_examples}).

\subsection{Data Retrieval Failures}

\paragraph{(a) Missing or Incomplete Retrieval.}
Agents frequently failed to retrieve existing laboratory values within the record. For example, cases were incorrectly flagged as \textit{``no recent potassium level found''} despite valid values being present. These failures likely stem from timestamp filtering errors or incomplete database queries.

\paragraph{(b) Incorrect Value Retrieval.}
In several other tasks, the agent retrieved incorrect data points, often due to sorting or temporal misalignment. This suggests the retrieval logic did not consistently select the most recent valid record, which happens when agents do not supply a \textit{``sort"} parameter to their query.

\end {multicols}{2}
\begin{tabhere}
\centering
\caption{Representative task examples of identified failure modes from EHR data. Each task tests specific reasoning capabilities such as data retrieval, aggregation, or threshold-based clinical decisions.}
\small
\begin{tabular}{p{2cm} p{4.1cm} p{2.5cm} p{3cm} p{4.0cm}}

\toprule
\textbf{Category} & \textbf{Example Instruction} & \textbf{True Value} & \textbf{Expected Output} & \textbf{Challenge Type} \\
\midrule
\textbf{Retrieval} &
\textit{``What's the latest PT result for patient \{{MRN\}}?''} &
Prothrombin Time (PT) = 16~s &
FINISH([16.0]) & Requires locating the most recent record among multiple timestamps. \\
\midrule
\textbf{Aggregation} &
\textit{``What is the average glucose of patient \{{MRN\}} over the last 24 hours?''} &
234.9~mg/dL &
FINISH([234.9]) &
Tests correct filtering by time window and inclusion of all measurements. \\
\midrule
\textbf{Threshold / Conditional} &
\textit{``Review elderly patient  \{{MRN\}}'s most recent potassium value. Order replacement if clinically low.''} &
3.0~mEq/L (low) &
POST  \{fhir\_api\_base\} MedicationRequest &
Evaluates multi-step reasoning with comparison to threshold range (3.5–5.5). \\
\bottomrule
\end{tabular}
\label{tab:failure_examples}
\end{tabhere}

\begin{multicols} {2}

\subsection{Aggregation and Calculation Errors}

In tasks involving empirical aggregation, such as \textit{``average glucose over last 24 hours"} , agents often generated incorrect results despite retrieving the correct input records. The errors typically arose from missing data within aggregation windows or arithmetic inconsistencies, indicating weaknesses in data inclusion logic and temporal boundary handling.

\subsection{Multi-step and Conditional Logic Failures}

Agents exhibited systematic weaknesses in applying conditional thresholds. For example in \textit{``potassium replacement"} tasks, 9 cases misclassified borderline values (2.0–2.2~mg/dL) as ``not low" even though the ground truth indicated replacement was warranted. This reflects discrepancies between hardcoded threshold values and context-dependent clinical decision logic.

While not strictly semantic errors, several outputs exhibited format inconsistencies (e.g., full ISO timestamps vs.\ date-only, or dictionary vs.\ tuple structure). Although these discrepancies do not impact correctness, they complicate downstream parsing and underscore the need for stricter schema enforcement in multi-agent medical reasoning pipelines.

\section{Proposed Approach}

Building on the identified failure modes, we propose a four-stage agentic framework to generate failure-targeted benchmark tasks that expose, quantify, and mitigate reasoning weakness in medical agents.

\subsection{Stage 1 - Scenario Identification Agent:} We first mine and extract failure-prone real-world instances of retrieval, aggregation, and threshold based conditional reasoning from electronic health records. Each case corresponds to a task type where prior models demonstrated systematic errors. From EHR observation data (50,000 records of 695 patients), we extracted scenarios matching each of the failure modes.

\paragraph{Data Synthesis:} Tasks are synthesized by extracting real patient scenarios (e.g., MRN, timestamps, demographics, lab results) from the EHR database and generating diverse natural language task instructions using extracted scenarios as a baseline. For example, a real patient record with potassium = 3.0 mEq/L at timestamp 2023-11-13 becomes the ground truth, while GPT-4o-mini generates variations like ``Review patient X's recent 
potassium" or ``Check if patient X needs electrolyte replacement." The real clinical data (e.g., lab results, timestamps, patient demographics) remains authentic and unchanged; only the instructional phrasing is synthesized. All synthesized task instructions are verified against the source EHR database before task inclusion.

\paragraph{Retrieval Tasks:} Identified patients with (a) sparse measurements ($\leq$3 total), (b) clustered timestamps (multiple readings within 2 hours), and (c) temporal edge cases (measurements at 23-25h boundaries). To ensure comprehensive coverage, we generated retrieval tasks for 11 distinct laboratory codes: CA (calcium), CL (chloride), CR (creatinine), PT (prothrombin time), TP (total protein), GLU (glucose), K (potassium), MG (magnesium), NA (sodium), HGB (hemoglobin), and HCT (hematocrit).

\paragraph{Aggregation Tasks:} Extracted recent valid 24-hour windows with $\geq$3 measurements, and generated tasks for glucose, potassium, and sodium laboratory codes.

\paragraph{Multi-step Conditional Tasks:} Sampled representative cases across 7 value ranges (critically low, low, borderline-low, normal, borderline-high, high, critically high) for laboratory codes with demographic-specific thresholds (hemoglobin, hematocrit, creatinine). Tasks required agents to (1) retrieve the lab value, (2) evaluate if abnormal based on age/gender-specific reference ranges, and (3) conditionally order appropriate interventions (e.g., ``Review most recent magnesium value for elderly male patient X. Order IV replacement if clinically low.''), testing both retrieval accuracy and clinical decision-making under ambiguous thresholds.

\subsection{Stage 2 - Task Generation Agent:} In this stage, large language models (e.g., GPT4o-mini, Claude 3.5) are employed to synthesize diverse natural language variants of each mined failure mode, expanding the benchmark’s linguistic and contextual diversity while maintaining clinical fidelity. Prompts are systematically varied in phrasing, context length, and conditional structure to ensure that the generated tasks reflect the complexity and variability of real-world clinical reasoning. This stage produces 200 synthetic tasks per failure mode, resulting in a balanced evaluation set grounded in real patient data.
Each task (see Fig 2 in Appendix) comprises four key components:
\vspace{-2mm}
\begin{itemize}
    \item Instruction – a task directive reflecting the clinical reasoning goal (e.g., “Compute average glucose over the last 24 hours”).
    \vspace{-1mm}
    \item Context – relevant patient demographic information required for reasoning (e.g., Male patient, age 73, routine glucose monitoring), and hospital-specific EHR system context like lab codes.
      \vspace{-1mm}
    \item Ground Truth – the correct response derived from validated EHR data (e.g.,  93.0 mg/dL).
      \vspace{-1mm}
    \item Evaluation Patient MRN – an anonymized identifier linking the task to a specific patient record (e.g., S6330912).
      \vspace{-1mm}
\end{itemize}

The agent response (see Fig 3 in Appendix) then reports the status of task completion, the result generated by the model, and a history summarizing the sequence of reasoning steps or actions performed during execution. Together, these components ensure structured, traceable benchmarking while preserving linguistic variability and clinical realism across all failure modes.

\subsection{Stage 3 - Quality Audit Agent: } Each generated task undergoes a multi-tier validation process to ensure clinical accuracy and consistency with the source electronic health record (EHR) data. Initially, tasks are automatically validated by aligning outputs with their originating entries, confirming that retrieved values, time windows, and thresholds correspond to real patient observations.

During the initial phase, human-in-the-loop (HITL) clinical experts were engaged to validate a representative subset of synthetic tasks, assessing each for factual relevancy and logical correctness. Expert review ensured that generated tasks accurately reflected clinical context, threshold interpretation, and reasoning consistency, thereby establishing a reliable baseline for automated validation.

As future work, we will scale validation using medical domain LLMs—MedGemma \cite{medgemma} and Med-PaLM \cite{medpalm}—as automated QA agents. These models will perform first-pass checks for factual relevancy, logical correctness, and threshold consistency, reducing the need for continuous human review. A dynamic confidence score-based routing mechanism will escalate low-confidence tasks for expert validation while auto-approving reliable outputs, creating an adaptive, scalable, and robust QA pipeline.



\subsection{Stage 4: AgentBench Evaluation}
We deploy GPT-4o-mini and Claude 3.5 Sonnet agents using MedAgentBench's evaluation framework \cite{agentbench}, executing tasks against a live EHR environment via FHIR-compliant API calls. Models operate in stateless mode (temperature=0, no conversation history) to prevent memorization effects. Each task is scored by exact-match comparison against ground-truth values computed from the EHR database.

\textbf{Limitation:} Exact-match accuracy does not verify reasoning quality - agents may reach correct outputs through flawed intermediate steps (e.g., incorrect temporal filtering that coincidentally retrieves the right value). Future work will incorporate step-wise correctness checks to distinguish sound clinical reasoning from fortuitous outcomes.

\end{multicols}
\begin{figure*}
			\centering
			\includegraphics[width=\linewidth]{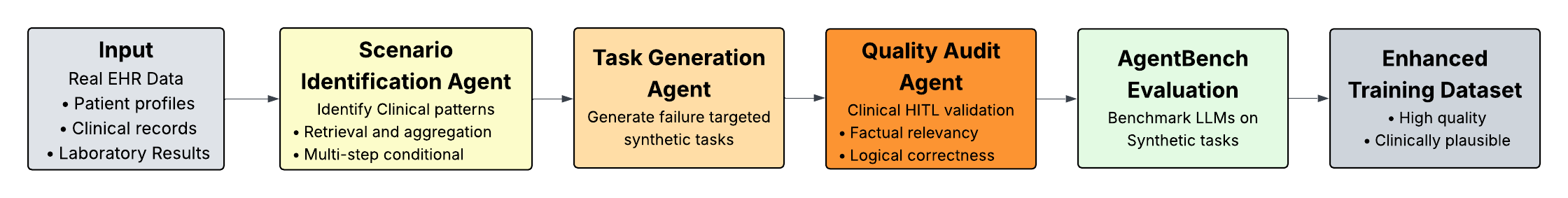}
            \vspace{-4mm}
			\caption{Agentic Task Generation \& Evaluation Pipeline}
			\label{fig:fullpictest1}
            \vspace{-4mm}
\end{figure*}

\begin{tabhere}
    \caption{Success Rate (SR, \%) with 200 tasks per category. }
    \centering
    \setlength{\tabcolsep}{9.1mm}{%
    \begin{tabular}{lcccc}
        \toprule
        \textbf{Category} & \textbf{\#Tasks} & \textbf{GPT-4o-mini} & \textbf{Claude-3.5} \\
        \midrule
        Retrieval & 200 & \textbf{100.0} & \textbf{100.0}  \\
        Aggregation (24h windows) & 200 & 28.0 & 64.0 \\
        Threshold / Conditional & 200 & 32.0 & 38.0 \\
        \bottomrule
    \end{tabular}%
    }
    \label{tab:sr_by_category}%
\end{tabhere}%

\begin{multicols}{2}
        
\section{Preliminary Results and Discussion}

Table 2 presents the projected Success Rate (SR; exact match) across three reasoning categories — data retrieval, aggregation, and threshold/conditional logic, each evaluated on an extended set of 200 synthetic tasks designed to target known failure scnearios. Both \textbf{GPT-4o-mini} and \textbf{Claude-3.5} achieved \textbf{100\% SR in retrieval tasks}, improving substantially from baseline scores of \textbf{42.5\%} and \textbf{54.5\%}, respectively, after the addition of an explicit \textit{“sort by date”} argument in the prompt. This demonstrates that large instruction-tuned models can achieve deterministic retrieval when task parameters are explicitly defined. However, performance declined sharply for \textbf{aggregation tasks}, which required temporal grouping over 24-hour window — \textbf{Claude-3.5} achieved \textbf{64\%} and \textbf{GPT-4o-mini 28\%}, indicating partial yet inconsistent handling of temporal and numerical reasoning. The weakest performance appeared in \textbf{threshold and conditional logic} tasks, with success rates of \textbf{38\% for Claude-3.5} and \textbf{32\% for GPT-4o-mini}, reflecting continuing limitations in multi-step logical evaluation and constraint chaining.

These quantitative results translate into tangible reliability concerns in clinical settings, where such reasoning behaviors underpin diagnostic interpretation and treatment decision support. The specific implications of each failure type are as follows:

\begin{itemize}
    \item \textbf{Data Retrieval Failures:} Missing or outdated laboratory results lead to incomplete patient profiles and hinder accurate clinical assessment. Such omissions may cause delayed recognition of abnormalities or incorrect assumptions about a patient’s current status.
    \item \textbf{Aggregation and Calculation Errors:} Inaccurate computation of averages or cumulative metrics, such as mean glucose or electrolyte trends, distorts longitudinal monitoring. These errors can result in improper medication titration, misinterpretation of disease progression, or failure to detect clinically significant changes.
    \item \textbf{Conditional and Threshold Misjudgments:} Rigid or incorrect application of threshold logic can directly compromise treatment safety. Misclassification of borderline or context-dependent values may prompt inappropriate therapeutic actions, including unwarranted interventions or omission of necessary care.
\end{itemize}

Together, these findings demonstrate that improving numerical reasoning, temporal aggregation, and context-sensitive logic through failure-targeted synthetic task generation is critical for developing clinically reliable and safety-aligned medical agents.

\section{Limitations}
Our benchmark currently includes 600 tasks across three 
failure modes, evaluated on only two models (GPT-4o-mini and Claude 3.5). We do not include baselines such as fine-tuned medical SLMs or ablation studies testing different retrieval strategies, limiting our ability to determine whether observed gaps stem from inherent LLM limitations or suboptimal tool use. Our exact-match accuracy metric does not capture reasoning quality—agents may produce correct outputs through incorrect intermediate steps. Without step-wise verification, we cannot distinguish between lucky guesses and sound clinical reasoning. Tasks are synthesized from a single institution's EHR formatted to FHIR R5 standards. Performance may not generalize to different EHR systems, coding standards, or clinical contexts outside the 11 lab codes tested.

\section{Future Work}
Future work will advance ART along these directions:

\begin{itemize}
    \item \textbf{Fine-grained Error Analysis}: Implement detailed error taxonomies distinguishing aggregation failures (window boundary errors, incomplete data inclusion, arithmetic mistakes) from threshold misjudgments (boundary misclassification, context-insensitive ranges, hallucinated values). Per-type metrics will enable targeted mitigation strategies.
    
    \item \textbf{Multi-turn Reasoning}: Add memory support to enable agents to cache prior values and maintain context across reasoning steps, allowing consistent multi-step decision-making rather than isolated single-call responses.
    
    \item \textbf{Robust Testing}: Evaluate temporal stability by varying time windows (24h → 36h) and threshold boundaries to assess model sensitivity to parameter shifts and edge-case scenarios.
    
    \item \textbf{Scalable Quality Assurance}: Integrate medical-domain LLMs (MedGemma \cite{medgemma}, Med-PaLM \cite{medpalm}) as automated QA agents with confidence-based routing—escalating low-confidence tasks ($\leq 0.6$) to clinical experts while auto-approving high-confidence outputs. Expand evaluation to additional models (GPT-5 \cite{openai2025gpt5}, medical SLMs \cite{kim2025small}) and deploy ART as a continuous monitoring framework for clinical decision support systems.
\end{itemize}

Together, these enhancements will enable a continuously improving, safety-aligned evaluation framework centered on failure-targeted synthetic data generation to strengthen clinical reasoning robustness in healthcare systems.

\bibliographystyle{unsrt}
\bibliography{Reference}

\appendix
\begin{figure*}
			\centering
            \caption{Task example for threshold-based failure mode}
			\includegraphics[width=0.9\linewidth]{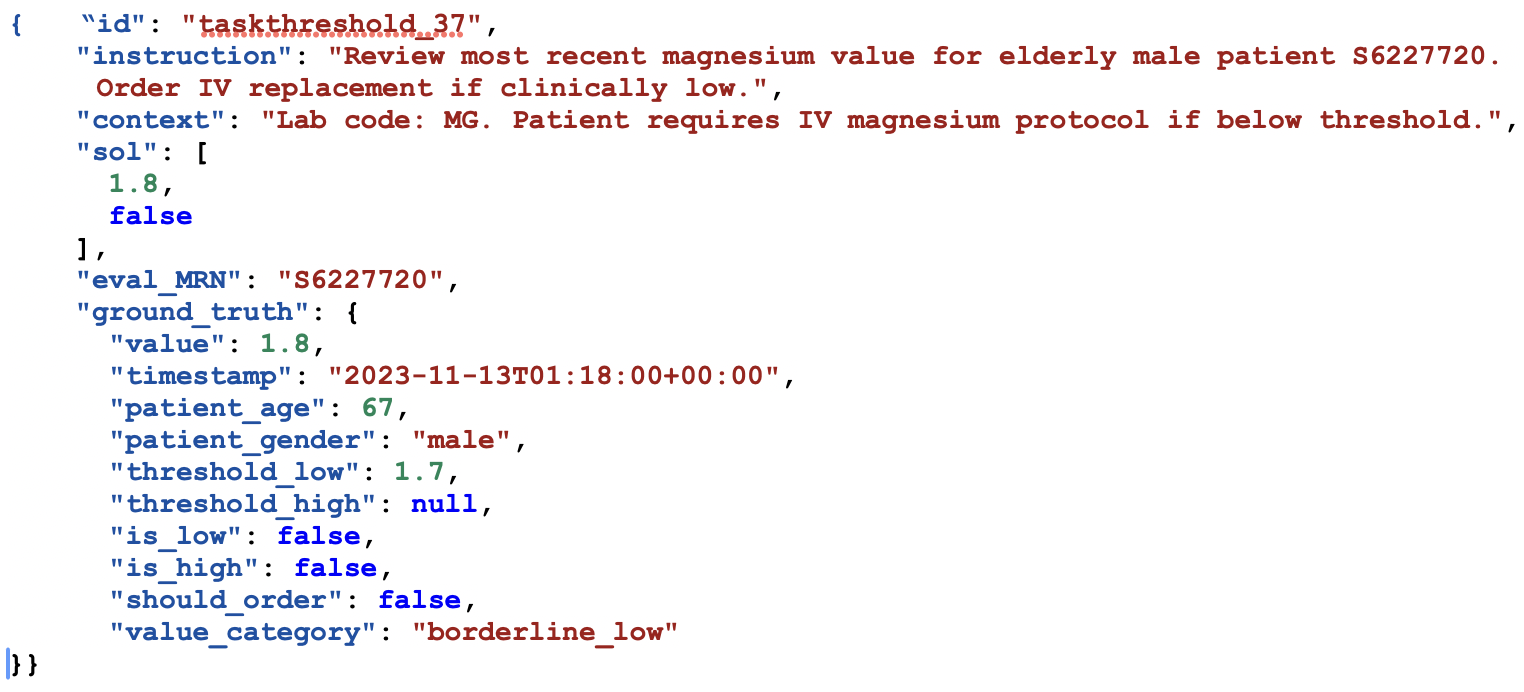}
            \vspace{-4mm}
			
			\label{fig:fullpictest1}
            \vspace{-8mm}
\end{figure*}
\begin{figure*}
			\centering
            \caption{Task Generator Agent response example}
			\includegraphics[width=0.9\linewidth]{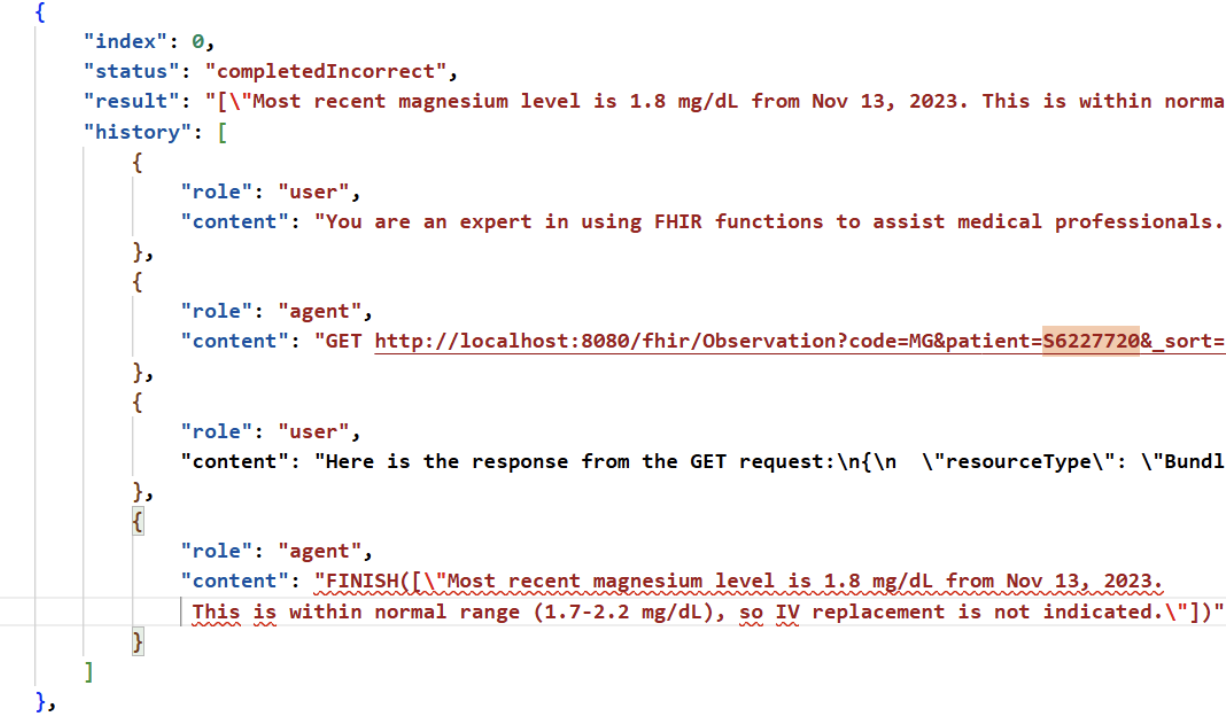}
            \vspace{-8mm}
			\label{fig:fullpictest1}
            \vspace{-4mm}
\end{figure*}
\end{multicols}{2}
\end{document}